\documentclass[sigconf,nonacm]{acmart}
\usepackage{algorithm}
\usepackage{algorithmic}
\usepackage{booktabs}
\usepackage{multirow}
\usepackage{xcolor}
\usepackage{makecell}
\AtBeginDocument{%
  }
\renewcommand\footnotetextcopyrightpermission[1]{}
\setcopyright{none}
\renewcommand\footnotetextcopyrightpermission[1]{}

\begin{document}

\title{Bi-Level Collaborative Learning for Few-Shot Scribble-Supervised Medical Image Segmentation}
\titlenote{%
The paper has been accepted for publication in the Proceedings of the
ACM International Conference on Multimedia (ACM MM 2026).
The final Version of Record will be available in the ACM Digital Library.
}
\author{Xiang-Xiang Su}
\email{sxxdyx0619@163.com}
\orcid{0009-0007-6104-8533}
\affiliation{%
  \institution{Fuzhou University}
  \city{Fuzhou}
  \country{China}
}

\author{Yufan Ye}
\orcid{0009-0006-9587-4190}
\email{241027164@fzu.edu.cn}
\affiliation{%
  \institution{Fuzhou University}
  \city{Fuzhou}
  \country{China}
}

\author{Yihang Zheng}
\orcid{0009-0000-7318-9762}
\email{2501027104@fzu.edu.cn}
\affiliation{%
  \institution{Fuzhou University}
  \city{Fuzhou}
  \country{China}
}

\author{Min Gan}
\orcid{0000-0002-2756-0054}
\email{aganmin@aliyun.com}
\affiliation{%
  \institution{Qingdao University}
  \city{Qingdao}
  \country{China}
}

\author{Guang-Yong Chen}
\correspondingauthor
\orcid{0000-0003-2088-9188}
\email{cgykeda@mail.ustc.edu.cn}
\affiliation{%
  \institution{Fuzhou University}
  \city{Fuzhou}
  \country{China}
}

\renewcommand{\shortauthors}{Su et al.}

\begin{abstract}
Scribble annotations offer an efficient alternative to costly pixel-wise labeling for medical image segmentation, yet in real clinical scenarios, scribble-annotated samples are often still limited, imposing the dual challenges of sparse supervision and annotated sample scarcity. These compounded constraints severely deprive models of the structural evidence needed for complete region recovery and precise boundary delineation. To break this bottleneck, we propose a bi-level collaborative learning framework for few-shot scribble-supervised medical image segmentation. Specifically, an upper-level learnable superpixel model is introduced to provide region-structural priors for lower-level segmentation, while superpixel-based region-wise pseudo-label propagation and a spatial-prior-guided filtering strategy are performed to generate reliable dense pseudo-labels for segmentation learning. Meanwhile, the anatomical semantics learned by the lower-level segmentation model under the guidance of the current superpixels are fed back to the upper level, further driving it to learn region-structural representations better aligned with the segmentation task. Through bidirectional interaction and collaborative learning between the upper and lower levels, the proposed framework significantly outperforms existing state-of-the-art scribble-supervised methods on the ACDC and Prostate datasets under the few-shot scribble-supervised setting.
\end{abstract}

\ccsdesc[500]{Computing methodologies~Image segmentation}
\ccsdesc[500]{Applied computing~Life and medical sciences}

\keywords{Medical image segmentation, Scribble-supervised learning, Superpixel, Bi-level Learning}
\begin{teaserfigure}
  \includegraphics[width=\textwidth]{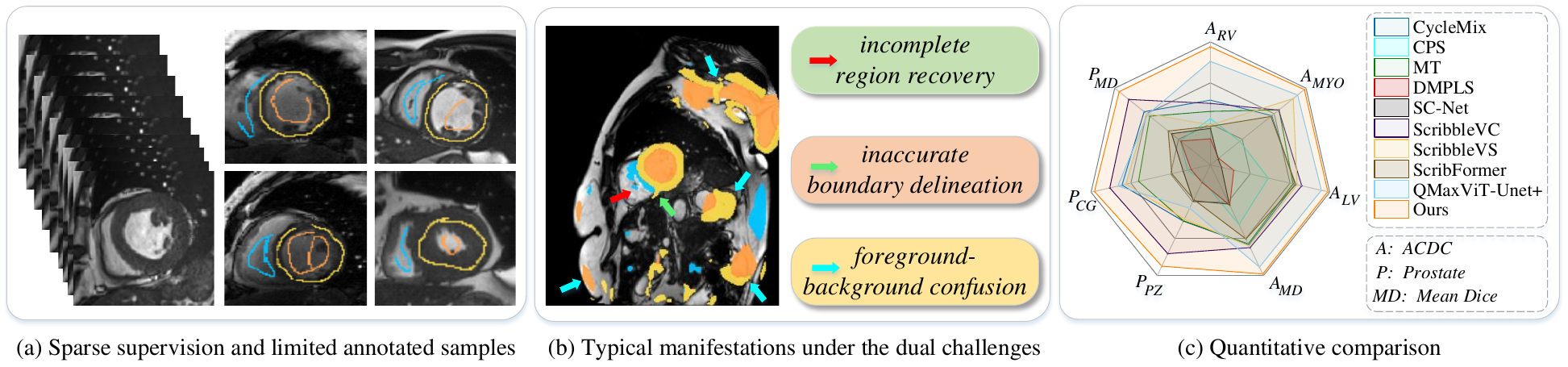}
  \caption{(a) Dual challenges in few-shot scribble-supervised segmentation: sparse supervision and limited annotated samples. (b) These dual challenges hinder the learning of stable anatomical representations, resulting in foreground-background confusion, incomplete region recovery, and inaccurate boundary delineation. (c) Quantitative and visual comparisons demonstrate the superior performance of the proposed method on benchmark datasets with five scribble-annotated training cases.}
  \label{fig:teaser}
\end{teaserfigure}


\maketitle

\section{Introduction}
Medical image segmentation~\cite{luo2022word,wu2024cross,li2024magic,lin2025gradient} plays a crucial role in disease diagnosis, treatment evaluation, and preoperative planning. However, obtaining high-quality pixel-level annotations typically requires experienced clinical experts to delineate anatomical structures on a pixel-by-pixel basis, which is both labor-intensive and time-consuming~\cite{tajbakhsh2020embracing}. To reduce this burden, scribble annotation has been increasingly adopted as an efficient and practical supervision form for weakly supervised medical image segmentation~\cite{ji2019scribble,PCE,VC}, since it only requires a small number of strokes within target regions to convey semantic category information. Yet, such sparse supervision provides neither complete object extent nor precise boundary information. Consequently, recovering reliable dense segmentation from extremely sparse scribble annotations remains highly challenging.

To compensate for the information missing from scribble annotations, existing scribble-based medical image segmentation methods typically employ data augmentation \cite{Cyclemix}, label propagation \cite{grady2006random,SC-Net}, pseudo-label generation \cite{WSL4,VC}, and consistency regularization \cite{VS,HELPNet} to transform sparse supervision into richer training signals. While these methods have achieved encouraging progress, they generally rely on access to abundant scribble-annotated data, enabling the model to learn anatomical priors that compensate for the sparsity of the supervision signal. 
However, reducing annotation density does not necessarily make labeled data abundant. In real clinical scenarios, obtaining scribble-annotated samples still requires considerable expert effort for case selection and data curation, and thus often remains limited, leading to the dual challenges of weak supervision and few-shot learning~\cite{fan2024bi,zhao2019data,zhang2024modelmix}.
These compounded constraints prevent the model from establishing stable anatomical representations, thereby exacerbating semantic confusion, incomplete region recovery, and inaccurate boundary delineation.


To address the above challenges, superpixels~\cite{achanta2012slic} serve as a promising region-level structural prior that can propagate sparse scribbles into denser region-level supervision and provide local structural cues for boundary recovery. However, existing superpixel-based methods~\cite{SC-Net,li2025scribble} mostly rely on region partitions derived by fixed algorithms from low-level features inherently present in the image. Such static priors can become unreliable when they are misaligned with anatomical semantics, especially under few-shot settings where structural evidence is limited. This raises a key question: \textit{how can we fully exploit the anatomically aligned structural priors encoded in superpixels to derive reliable dense supervisory signals for few-shot scribble-supervised segmentation?}

Motivated by this, we propose a bi-level collaborative learning framework for few-shot scribble-supervised medical image segmentation, termed \textbf{BiSCL}, which couples a learnable superpixel network with a segmentation network. Specifically, the upper level introduces a learnable superpixel model to continuously provide region-structural and boundary priors for lower-level segmentation. At the lower level, these superpixel priors are exploited for region-wise pseudo-label propagation and a spatial-prior-guided filtering strategy is performed to generate reliable dense pseudo-labels for segmentation learning. Meanwhile, the anatomical semantics learned by the segmentation model are fed back to the upper level, further driving the superpixel model to learn region-structural representations that are better aligned with the segmentation objective. Through such bidirectional collaboration and mutual refinement, the proposed BiSCL achieves more reliable region recovery and boundary delineation in few-shot scribble-supervised medical image segmentation.

The main contributions of this work are summarized as follows:
\begin{itemize}
\item We propose a bi-level collaborative learning framework that couples superpixel modeling with segmentation for few-shot scribble-supervised medical image segmentation, where bidirectional interaction between the upper and lower levels jointly improves region structure modeling and segmentation performance.
\item We introduce a spatial-prior-guided adaptive filtering strategy to suppress incorrect pseudo-label propagation across superpixels, thereby reducing error accumulation.
\item We develop an effective bilevel optimization scheme to support collaborative learning between the superpixel and segmentation networks, and extensive experiments show that the proposed method outperforms existing methods.
\end{itemize}

\begin{figure*}[htp]
    \centering
    \includegraphics[width=1.0\linewidth]{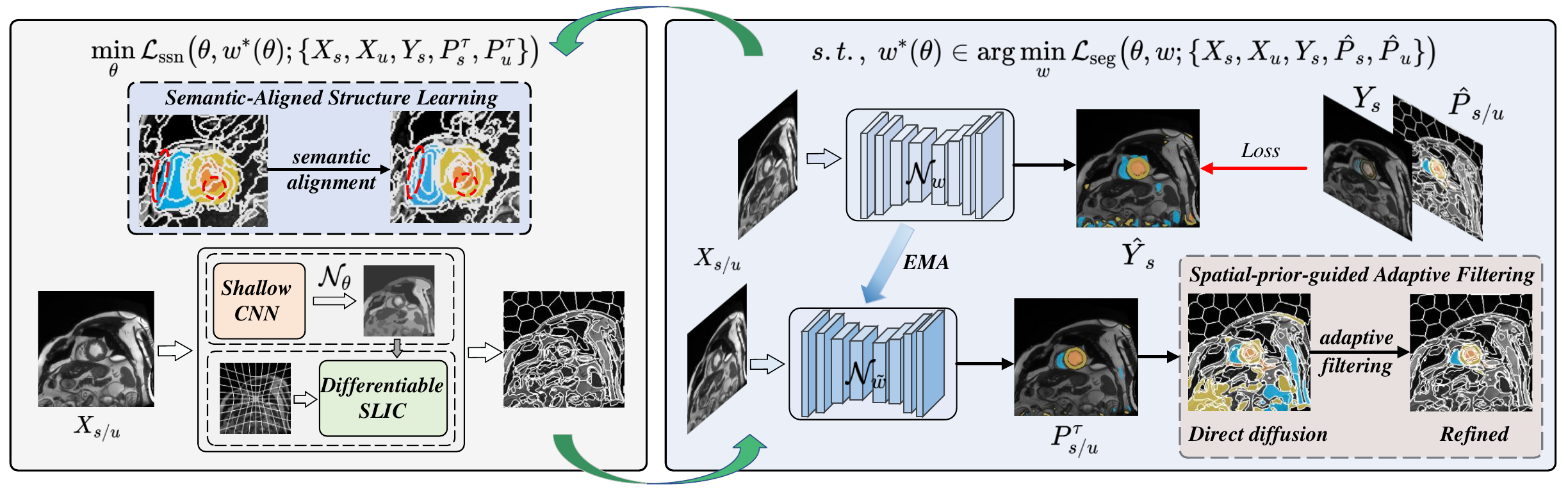}
    \caption{An overview of the proposed BiSCL framework.}
    \label{fig:framework}
\end{figure*}
\section{Related Work}
\subsection{Scribble-based Medical Image Segmentation}
Scribble annotation can substantially reduce the labeling cost of medical image segmentation, but its extremely sparse supervision makes reliable dense prediction highly challenging. Early studies typically employ partial cross-entropy~\cite{PCE} to supervise only annotated pixels, while more recent methods focus on transforming sparse scribbles into denser and more reliable supervision. Existing approaches can be roughly categorized into three lines. The first focuses on pseudo-label generation and consistency learning. For example, DMPLS~\cite{WSL4} generates auxiliary pseudo-labels with a dual-decoder architecture, while ScribbleVS~\cite{VS} further improves pseudo-label reliability under the Mean Teacher framework. The second line improves robustness through data augmentation and regularization. Representative methods include CycleMix~\cite{Cyclemix}, which introduces image-mixing and cycle consistency, and HELPNet~\cite{HELPNet}, which combines hierarchical perturbation consistency with entropy-guided regularization. The third line addresses missing structural information by introducing stronger semantic modeling, explicit structural priors, or more advanced architectures. For instance, ScribbleVC~\cite{VC} strengthens image--class correspondence through visual--class embeddings, ScribFormer~\cite{li2024scribformer} explores a CNN--Transformer hybrid architecture, and QMaxViT-Unet+~\cite{QMaxvit} further improves feature representation and boundary perception with a transformer-based backbone and edge enhancement design. Despite promising progress, these methods generally rely on sufficient scribble annotations to learn stable anatomical representations. When annotated samples are limited, insufficient structural evidence hampers complete region recovery and precise boundary delineation, leading to substantial performance degradation.
\subsection{Superpixels in Medical Image Segmentation}
Superpixel segmentation over-segments an image into spatially contiguous and visually similar regions, providing a boundary-aware and computationally efficient mid-level representation. Traditional methods typically rely on hand-crafted features with iterative clustering or energy optimization, such as SLIC~\cite{achanta2012slic,SLIC} and SEEDS~\cite{SEEDS}, while recent deep superpixel methods have increasingly explored learnable superpixel generation~\cite{jampani2018superpixel,yang2020superpixel,xu2023esnet}. Benefiting from their ability to preserve low-level structures and boundary cues, superpixels have been widely used in medical image segmentation for regional consistency modeling~\cite{li2021superpixel}, representation learning~\cite{wang2022separated}, and supervision refinement~\cite{thompson2022pseudo}. In weakly supervised settings, they further serve as effective regional priors to compensate for sparse annotations. For example, Li et al.~\cite{li2025scribble} combine multi-scale superpixels with deep features for dynamic pseudo-label generation. SP\textsuperscript{3}~\cite{li2024sp} propagates scribble labels within superpixels and refines pseudo-labels with dynamic thresholding and superpixel-level uncertainty. SC-Net~\cite{SC-Net} combines superpixel-guided scribble walking with class-wise contrastive regularization to expand sparse scribbles into unlabeled regions. However, these methods predominantly treat superpixels as static priors. Once superpixel partitions are misaligned with target semantics, pseudo-label noise may be introduced and progressively amplified, especially under sparse supervision and annotated sample scarcity.
\section{Method}
\subsection{Preliminaries}
Let $\mathcal{D}_s=\{(X_s,Y_s)\}$ and $\mathcal{D}_u=\{X_u\}$ denote the scribble-labeled and unlabeled datasets, respectively, where $X_s$ and $X_u$ are input images and $Y_s$ is the scribble annotation associated with $X_s$. Given an arbitrary segmentation architecture, we denote the student and teacher models as $\mathcal{N}_w(\cdot)$ and $\mathcal{N}_{\tilde{w}}(\cdot)$, respectively, where $w$ denotes the parameters of the student model and the teacher model is updated via the exponential moving average (EMA) of the student model. We further denote the superpixel network by $\mathcal{N}_{\theta}(\cdot)$, with $\theta$ representing its parameters. For labeled and unlabeled inputs, the teacher model produces confidence-filtered pseudo-labels $P_s^{\tau}$ and $P_u^{\tau}$, respectively. Subsequently, using the superpixel maps generated by $\mathcal{N}_{\theta}$, we obtain refined pseudo-labels $\hat{P}_s$ and $\hat{P}_u$ by applying the proposed adaptive diffusion strategy to $P_s$ and $P_u$.
\subsection{Bilevel Framework}
In real-world medical applications, the scarcity of dense annotations and limited labeled samples often coexist, posing a coupled challenge of weak supervision and few-shot learning. Sparse scribble supervision lacks complete region and boundary-structure information, making the model prone to unstable local predictions, inaccurate boundary delineation, and pseudo-label noise accumulation. Meanwhile, the limited scribble-annotated samples are insufficient to support adequate structural induction, further restricting the model’s ability to capture anatomical semantics and morpho-structural characteristics of the target.

To address the dual challenges of few-shot scribble segmentation, we introduce superpixels to provide more informative region-level supervision and boundary cues. Unlike existing paradigms that directly treat superpixels as static external priors for supervision expansion, we seek to make superpixel partitioning dynamically evolve with the optimization of the segmentation objective, thereby yielding structural priors that are better aligned with the target anatomical semantics. Therefore, we propose BiSCL, a bilevel collaborative learning framework between superpixels and segmentation, and formulate it as the following bilevel optimization problem:
\begin{equation}\label{BLO}
\begin{split}
&\min_{\theta}\mathcal{L}_{\text{ssn}}\big(\theta,w^*(\theta);\{X_s,X_u,Y_s,P_s^{\tau},P_u^{\tau}\}\big)\\
& s.t.,\ w^*(\theta)\in \arg\min\limits_{w}\mathcal{L}_{\text{seg}}\big(\theta,w;\{X_s,X_u,Y_s,\hat{P}_s,\hat{P}_u\}\big)
\end{split}
\end{equation}
where the lower-level task aims to maximize segmentation performance under the region and boundary priors provided by the upper-level superpixel network, while feeding back the optimized anatomical semantics learned under the current superpixel guidance to the upper level. This in turn drives the superpixel network to learn structural representations that are better aligned with the lower-level segmentation task, thereby forming a collaborative learning loop in which the upper level provides structural priors to the lower level, and the lower level in turn offers semantic correction to the upper level.


\subsection{Lower-Level Segmentation Task}
The core of lower-level learning lies in effectively leveraging the region-level structural priors provided by the upper-level superpixel network to construct reliable dense supervision under limited scribble annotations. To achieve this, we adopt the Mean Teacher framework, expand the teacher’s high-confidence pseudo-labels via superpixels, and further develop a spatial-prior-guided adaptive filtering strategy to suppress erroneous propagation.

Specifically, for scribble-annotated samples, we compute the partial cross-entropy loss as
\begin{equation}
\mathcal{L}_{\mathrm{pCE}}=-\sum_{i\in\Omega_s}\sum_{c=1}^{C}
(Y_s)_{i,c}\log (\hat{Y}_s)_{i,c},
\label{eq:pce}
\end{equation}
where $\hat{Y}_s=\text{Softmax}(\mathcal{N}_w(X_s))$ is the prediction of the student network, $\Omega_{s}$ denotes the set of scribble-labeled pixels, and $Y_s$ is the corresponding supervision label.

To obtain denser and more reliable supervision signals, we first compute the teacher probability map $Y_{s/u}^t=\text{Softmax}(\mathcal{N}_{\tilde{w}}(X_{s/u}))$, from which the pseudo-labels are derived as $P_{s/u}=\arg\max(Y_{s/u}^t)$. We further define the corresponding high-confidence mask as
\begin{equation}
    M_{s/u}^{\tau}=\mathbb{I}\big(\max_c (Y_{s/u}^t)_{i,c}>\tau\big),
\end{equation}
where $\mathbb{I}(\cdot)$ denotes the indicator function and $\tau$ is the confidence threshold. Based on the confidence mask, we construct the high-confidence seed labels as
\begin{equation}
    P_{s/u}^{\tau}=M_{s/u}^{\tau}\odot P_{s/u},
\end{equation}
where pixels with low confidence are suppressed and thus excluded from subsequent diffusion.

Based on the high-confidence seed labels, we further perform superpixel-guided label diffusion to expand sparse supervision into region-level pseudo-labels. Let $S(i)\in\{1,\dots,K\}$ denote the superpixel index of pixel $i$, and let $\Omega_k=\{i\mid S(i)=k\}$ denote the set of pixels belonging to the $k$-th superpixel. For each superpixel $k$, we first define its dominant seed class as
\begin{equation}
\hat{c}_k=\underset{c\in\{1,\dots,C-1\}}{\arg\max}
\sum_{j\in\Omega_k}\mathbb{I}\big(P_{s/u}^{\tau}(j)=c\big).
\label{eq:dominant_class}
\end{equation}
Then, for any $i\in\Omega_k$, the diffused pseudo-labels are defined as
\begin{equation}
P_{s/u}^{sp}(i)=
\begin{cases}
\hat{c}_k, & \exists\, j\in\Omega_k,\ P_{s/u}^{\tau}(j)>0,\\
0, & \text{otherwise},
\end{cases}
\label{eq:sp_diffusion}
\end{equation}
That is, for each superpixel, if high-confidence foreground seeds are present, all pixels within that superpixel are assigned to the dominant seed class; otherwise, the superpixel remains unlabeled. In this way, sparse high-confidence supervision can be expanded to structurally consistent regions, yielding denser pseudo-labels for subsequent segmentation learning.

However, directly using $P^{sp}_{s/u}$ may cause over-diffusion, especially when superpixels fail to align with semantic boundaries or the teacher produces inaccurate predictions. Under sparse scribble supervision and limited annotated data, such erroneous propagation can further amplify pseudo-label noise, ultimately degrading segmentation performance. As shown in Fig.~\ref{fig:framework}, a small false-positive prediction in the background may be expanded to an entire large superpixel, substantially enlarging the error region. To suppress this effect, we introduce a spatial-prior-guided adaptive expansion-ratio filtering strategy. Specifically, for each superpixel $k$, we define its expansion ratio as
\begin{equation}
r_k=
\frac{|\Omega_k|}
{\sum_{j\in\Omega_k}\mathbb{I}\big(P_{s/u}^{\tau}(j)=\hat{c}_k\big)},
\label{eq:expansion_ratio}
\end{equation}
which measures how much the dominant seed support is enlarged after superpixel diffusion. Based on the empirical prior that the target region is more likely to be located near the image center, the maximum allowable expansion ratio is adjusted according to the distance between each superpixel centroid and the image center. Specifically, let $\mathbf{c}_{k}$ denote the centroid of the $k$-th superpixel and let $\mathbf{c}_{0}$ denote the image center. The normalized distance between them is defined as
\begin{equation}
d_k=\frac{\|\mathbf{c}_k-\mathbf{c}_0\|_2}{d_{\max}},
\end{equation}
Based on this distance, the maximum allowable expansion ratio for superpixel $k$ is defined as
\begin{equation}
\gamma_k=1+(\gamma_{\mathrm{center}}-1)
\exp\left(-\frac{d_k^2}{2\sigma^2}\right),
\end{equation}
where $\gamma_{\mathrm{center}}\ge1$ denotes the maximum expansion ratio allowed near the image center and $\sigma$ controls the decay rate. The diffused pseudo-labels within superpixel $k$ are retained only when $r_k\le \gamma_k$; otherwise, the corresponding diffusion result is discarded. Accordingly, the final filtered pseudo-labels are defined as
\begin{equation}
    \hat{P}_{s/u}(i)=
    \begin{cases}
    P_{s/u}^{sp}(i), & r_k\le \gamma_k,\quad i\in\Omega_k,\\
    0, & \text{otherwise}.
    \end{cases}
    \label{eq:filtered_pseudo_label}
\end{equation}

The pseudo-label supervision term is then defined as
\begin{equation}
\mathcal{L}_{\mathrm{pseudo}}
=0.5*\big(\mathcal{L}_{\mathrm{CE}}(\hat{Y}_{s/u},\hat{P}_{s/u}\big)+\mathcal{L}_{\mathrm{Dice}}(\hat{Y}_{s/u},\hat{P}_{s/u})\big).
\end{equation}
The final lower-level objective is written as
\begin{equation}
\mathcal{L}_{\mathrm{seg}}
=
\mathcal{L}_{\mathrm{pCE}}
+
\lambda(t)\mathcal{L}_{\mathrm{pseudo}},
\label{eq:lseg}
\end{equation}
where $\lambda(t) = \text{exp}(-5(1-t/T)^2)$ is a Gaussian ramp-up schedule, with $t$ being the current training step and $T$ the maximum training step.

\subsection{Upper-Level Superpixel Task}
The upper-level task aims to learn structural priors tailored to the lower-level segmentation task, enabling superpixels to preserve sensitivity to the original image boundaries while forming region representations that are more aligned with the anatomical semantics of the segmentation target.

To this end, we adopt a differentiable Superpixel Sampling Network (SSN)~\cite{jampani2018superpixel} as the upper-level structural modeler. Specifically, the network represents the pixel-to-superpixel assignment as a learnable soft association matrix and obtains superpixel partitions through iterative assignment in a joint feature space. For superpixel initialization, instead of adopting a uniform grid, we introduce a spatial-prior-guided non-uniform strategy that allocates superpixel centers more densely in the central region and more sparsely toward the image periphery. Given a pixel location $p$, we first compute its normalized radial distance to the image center
\begin{equation}
r(p)=\frac{\|p-c_0\|_2}{r_{\max}},
\label{eq:radial_norm}
\end{equation}
where $r_{\max}$ is the maximum distance from the image center to the image corners. We then apply an inverse radial transform
\begin{equation}
\tilde{r}(p)=r(p)^{1/\gamma}, \qquad \gamma \ge 1,
\label{eq:radial_warp}
\end{equation}
where $\gamma$ controls the strength of the center-dense reallocation. The warped coordinates are subsequently quantized on a uniform seed grid to generate the initial superpixel labels. In this way, the proposed strategy allocates denser superpixel seeds in central regions and sparser ones toward the periphery, thereby concentrating more structural capacity on potential target areas, reducing sensitivity to the prescribed number of superpixels, and providing a more suitable structural starting point for subsequent semantic alignment.

To enable the superpixel network to better balance structural boundary preservation and semantic alignment, we construct a joint optimization objective consisting of low-level feature reconstruction, semantic reconstruction, and compactness regularization. Specifically, to preserve the sensitivity of superpixels to true image boundaries, we first introduce a low-level feature reconstruction loss. Let $R_{\mathrm{low}}$ denote the low-level image attributes, including intensity, gradients, and edge responses. Given the pixel--superpixel association matrix $Q$ predicted by $\mathcal{N}_{\theta}$, we reconstruct these attributes through pixel--superpixel--pixel projection:
\begin{equation}
R_{\mathrm{low}}^{*}=\tilde{Q}\hat{Q}^{\top}R_{\mathrm{low}},
\end{equation}
where $\hat{Q}$ and $\tilde{Q}$ denote the column-normalized and row-normalized association matrices, respectively. The corresponding reconstruction loss is defined as
\begin{equation}
\mathcal{L}_{\mathrm{low}}
=
\frac{1}{N}\sum_{i=1}^{N}\omega_i
\left\|
R_{\mathrm{low},i}-R_{\mathrm{low},i}^{*}
\right\|_1,
\label{eq:lfeat}
\end{equation}
where $w_i = 1 + \beta \ \mathrm{norm}(|\nabla I_i|)$ is the boundary-aware weight for pixel $i$, $\mathrm{norm}(\cdot)$ denotes per-image normalization, and $\beta$ controls the strength of boundary emphasis. This term encourages the superpixel network to capture local structures consistent with the original image boundaries and prevents it from being prematurely biased by unreliable semantic feedback from the lower level. Meanwhile, we retain the standard compactness regularization to enforce spatial regularity and suppress overly fragmented or elongated superpixel partitions:
\begin{equation}
\mathcal{L}_{\mathrm{compact}}
=
\left\|I_{xy}-\bar{I}_{xy}\right\|_2,
\label{eq:lcompact}
\end{equation}
where $I_{xy}$ and $\bar{I}_{xy}$ denote the original and reconstructed spatial coordinates, respectively.

To enable the upper-level superpixels to provide structural priors that are better aligned with the semantics of the lower-level segmentation task, we feed the task semantics learned by the lower-level segmentation model back to the upper level, thereby correcting its structural representation. Specifically, we construct a semantic reconstruction constraint using confidence-filtered teacher pseudo-labels, and further incorporate masked semantic reconstruction based on sparse yet reliable scribble annotations. The complete semantic reconstruction loss is defined as follows:
\begin{equation}
\begin{split}
    \mathcal{L}_{\mathrm{sem}}=&\frac{1}{2}\big(\mathrm{CE}\big(M_s\odot Y_s,\ M_s\odot (\tilde{Q}\hat{Q}^{\top}Y_s)\big)+\mathrm{CE}(P_s^{\tau},\tilde{Q}\hat{Q}^{\top}P_s^{\tau})\big)\\
    &+\mathrm{CE}(P_u^{\tau},\tilde{Q}\hat{Q}^{\top}P_u^{\tau}),
\end{split}
\end{equation}
where $M_s$ denotes the binary mask of scribble-annotated pixels.
The final upper-level objective is thus formulated as
\begin{equation}
\mathcal{L}_{\mathrm{ssn}}
=
\lambda_{\mathrm{low}}\mathcal{L}_{\mathrm{low}}
+
\lambda_{\mathrm{sem}}\mathcal{L}_{\mathrm{sem}}
+
\lambda_{\mathrm{c}}\mathcal{L}_{\mathrm{compact}},
\label{eq:lssn}
\end{equation}
where $\lambda_{\mathrm{low}}$, $\lambda_{\mathrm{sem}}$ and $\lambda_{\mathrm{c}}$ are the trade-off coefficients.

Through this upper-level optimization, the superpixel network is driven not only by low-level image evidence but also by task semantics progressively acquired by the lower-level segmentation. As a result, the generated superpixels gradually evolve from generic boundary-aware partitions to segmentation-oriented structural priors, which in turn provide more reliable structural support for pseudo-label diffusion and boundary modeling in the lower level.

\subsection{Bilevel Training Strategy}
This section presents the training strategy of the proposed bi-level framework. Since the pseudo-labels generated by the segmentation model are often noisy in the early stage of training, especially under few-shot scribble supervision, directly optimizing the segmentation and superpixel models from scratch may cause unstable early predictions to mislead the superpixel model into learning region representations with semantic bias, which can further amplify erroneous supervision during subsequent pseudo-label diffusion. To mitigate error accumulation, we pretrain the segmentation model and the superpixel model separately before bi-level collaborative training: the former is pretrained with the Mean Teacher strategy, while the latter is trained independently, without semantic feedback, to first capture low-level structural priors of the image.

After pretraining, we further develop a simple yet effective bi-level training strategy to collaboratively optimize the segmentation model and the superpixel model. Given the nested structure of the bilevel formulation \eqref{BLO} and the coupling relationship between the upper- and lower-level tasks, we first minimize the lower-level problem with fixed upper-level variable $\theta$:
\begin{equation}\label{eq-w}
    w\leftarrow w - \eta_{w}\nabla_w\mathcal{L}_{\mathrm{seg}}(w,\theta),
\end{equation}
where $\eta_{w}$ is the step size of student model $\mathcal{N}_w$, and the teacher model $\mathcal{N}_{\tilde{w}}$ is updated by EMA:
\begin{equation}\label{eq-ema}
    \tilde{w}\leftarrow \alpha\tilde{w} + (1-\alpha)w,
\end{equation}
where $\alpha$ denotes the moving average coefficient. After iteratively updating Eqs.~\eqref{eq-w}--\eqref{eq-ema} for $N_l$ steps, we obtain an approximate lower-level optimum $w^*(\theta)$, representing the best segmentation model parameters learned under the guidance of the current superpixel model. This solution is then fed back to the upper-level objective to refine the superpixel model toward learning structural representations that are better aligned with the semantic requirements of the lower-level segmentation task. Since $w^*(\theta)$ is implicitly dependent on $\theta$, updating the upper-level parameters in standard bilevel optimization requires computing the hypergradient of the upper-level objective:
\begin{equation}
    \nabla_{\theta}\mathcal{L}_{\mathrm{ssn}}(\theta,w^*(\theta)) + \big(\frac{dw^*(\theta)}{d\theta}\big)^{\text T}\nabla_{w}\mathcal{L}_{\mathrm{ssn}}(\theta,w^*(\theta)).
\end{equation}
However, estimating the Jacobian matrix $\frac{d w^*(\theta)}{d\theta}$ is typically computationally expensive, as it involves the evaluation of higher-order derivatives~\cite{liu2021investigating,zhang2024introduction}. 
Although neglecting this implicit gradient term may introduce bias, using only the upper-level partial gradient often works well in practice for deep bilevel problems~\cite{jiang2025beyond,jiang2025efficient}, while being substantially more efficient. Accordingly, we update the upper-level parameters as follows:
\begin{equation}\label{eq-theta}
\theta \leftarrow \theta - \eta_{\theta}\nabla_{\theta}\mathcal{L}_{\mathrm{ssn}}(\theta, w^*(\theta)).
\end{equation}
In classical bilevel optimization, each update of the upper-level variable typically requires re-solving the lower-level problem to obtain the corresponding lower-level optimum. To improve the reuse efficiency of lower-level semantic feedback, we perform $N_u$ iterations of \eqref{eq-theta} while fixing the current approximate lower-level solution $w^*(\theta)$, enabling the superpixel model to provide structural priors that are better aligned with subsequent lower-level segmentation updates. The complete training procedure is summarized in Algorithm~\ref{alg:1}.
\begin{algorithm}[H]
	\caption{Bi-level Collaborative Training Strategy}\label{alg:1}
	\begin{algorithmic}[1]
		\REQUIRE The initial parameters $w_0$, $\theta_0$, step-sizes $\eta_{\theta}$, $\eta_{w}$, lower-loop steps $N_l$, upper-loop steps $N_u$, the maximum number of iterations $T$, moving average coefficient $\alpha$, $t\leftarrow1$, $\hat{\theta}\leftarrow\theta_0$, $\tilde{w}\leftarrow w_0$.
		\ENSURE The optimal parameters $w^*$, $\theta^*$.
        \WHILE{$t<T$}
        \STATE \% Update the lower-level parameter.
        \FOR{$k=0:N_l-1$}
        \STATE $w_{k+1} = w_k - \eta_{w}\nabla_w\mathcal{L}_{\mathrm{seg}}(w_k,\hat{\theta})$.
        \STATE $\tilde{w}\leftarrow \alpha\tilde{w} + (1-\alpha)w_{k+1}$ \% Update $\mathcal{N}_{\tilde{w}}$ by EMA.
        \STATE $t=t+1$.
        \ENDFOR
        \STATE $\hat{w}\leftarrow w_{N_l}$, $\theta_0\leftarrow\hat{\theta}$.
        \STATE \% Update the upper-level parameter.
        \FOR{$k=0:N_u-1$}
        \STATE $\theta_{k+1} = \theta_k - \eta_{\theta}\nabla_{\theta}\mathcal{L}_{\mathrm{ssn}}(\hat{w},\theta_k)$.
        \ENDFOR
        \STATE $\hat{\theta}\leftarrow\theta_{N_u}$, $w_0\leftarrow\hat{w}$.
        \ENDWHILE
        \STATE Return $w^* \leftarrow \hat{w}$, $\theta^* \leftarrow \hat{\theta}$.
	\end{algorithmic}
\end{algorithm}
\begin{figure*}[htp]
    \centering
    \includegraphics[width=1\linewidth]{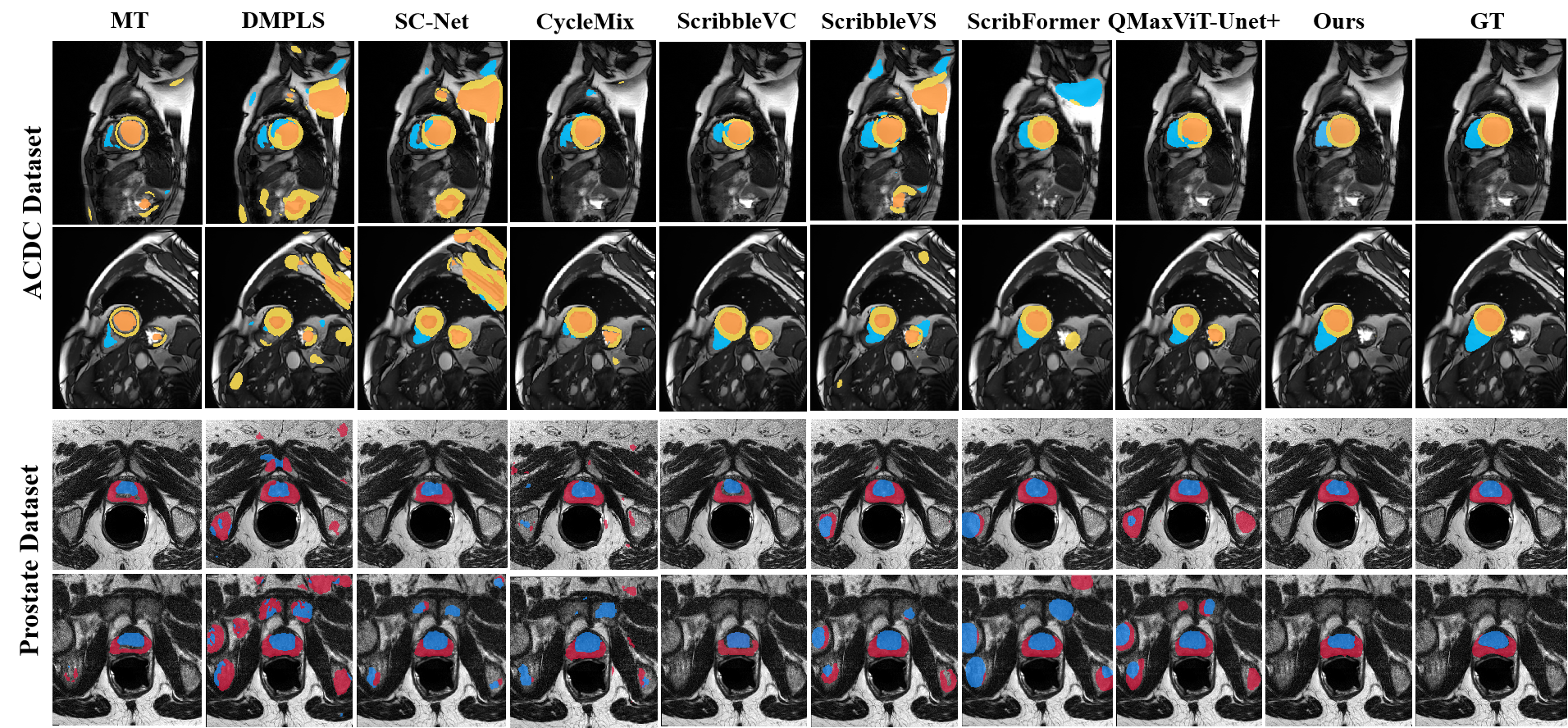}
    \caption{Qualitative comparison of our method with state-of-the-art methods on the ACDC and Prostate test sets under the setting of five scribble-annotated training cases.}
    \label{fig:acdc}
\end{figure*}
\begin{table*}[t]
\caption{Quantitative comparison of Dice and HD95 with baseline methods on the ACDC test set under the setting of five scribble-annotated training cases. The best and second-best results are highlighted in red and blue, respectively.}
\label{tab:acdc}
\centering
\small
\setlength{\tabcolsep}{4.8pt}
\begin{tabular}{lcccccccc}
\toprule
\multirow{2}{*}{Methods} & \multicolumn{4}{c}{Dice $\uparrow$} & \multicolumn{4}{c}{HD95 (mm) $\downarrow$} \\
\cmidrule(lr){2-5} \cmidrule(lr){6-9}
 & RV & MYO & LV & Mean & RV & MYO & LV & Mean \\
\midrule
PCE\cite{PCE}
& $0.261 \pm 0.160$ & $0.183 \pm 0.034$ & $0.311 \pm 0.143$ & $0.251 \pm 0.084$
& $114.26 \pm 10.48$ & $108.96 \pm 12.02$ & $115.25 \pm 14.28$ & $112.82 \pm 11.26$ \\

Cutout\cite{Cutout}
& $0.235 \pm 0.207$ & $0.253 \pm 0.067$ & $0.440 \pm 0.222$ & $0.309 \pm 0.140$
& $136.36 \pm 24.63$ & $113.39 \pm 11.04$ & $122.28 \pm 15.83$ & $124.01 \pm 11.89$ \\

CutMix\cite{Cutmix}
& $0.262 \pm 0.195$ & $0.324 \pm 0.091$ & $0.233 \pm 0.142$ & $0.273 \pm 0.120$
& $115.87 \pm 13.25$ & $116.64 \pm 13.28$ & $126.63 \pm 14.40$ & $119.72 \pm 11.93$ \\

CycleMix\cite{Cyclemix}
& $0.516 \pm 0.239$ & $0.604 \pm 0.141$ & $0.697 \pm 0.183$ & $0.606 \pm 0.166$
& $45.20 \pm 20.47$ & $50.52 \pm 18.61$ & $49.32 \pm 32.85$ & $48.35 \pm 18.26$ \\

CPS\cite{CPS}
& $0.426 \pm 0.171$ & $0.394 \pm 0.073$ & $0.540 \pm 0.170$ & $0.453 \pm 0.107$
& $104.16 \pm 15.48$ & $99.82 \pm 11.75$ & $101.35 \pm 14.99$ & $101.78 \pm 11.93$ \\

MT\cite{MT}
& $0.461 \pm 0.147$ & $0.652 \pm 0.103$ & $0.685 \pm 0.147$ & $0.599 \pm 0.106$
& {\color{blue}$22.44 \pm 30.11$} & $65.28 \pm 28.80$ & $73.63 \pm 35.87$ & $53.78 \pm 25.43$ \\

DMPLS\cite{WSL4}
& $0.327 \pm 0.216$ & $0.254 \pm 0.051$ & $0.337 \pm 0.190$ & $0.306 \pm 0.119$
& $82.76 \pm 25.25$ & $106.74 \pm 11.38$ & $106.37 \pm 18.41$ & $98.62 \pm 14.43$ \\

SC-Net\cite{SC-Net}
& $0.378 \pm 0.228$ & $0.250 \pm 0.082$ & $0.265 \pm 0.148$ & $0.298 \pm 0.116$
& $82.64 \pm 25.99$ & $108.22 \pm 13.62$ & $110.58 \pm 15.30$ & $100.48 \pm 15.08$ \\

ScribbleVC\cite{VC}
& $0.500 \pm 0.244$ & $0.650 \pm 0.138$ & $0.736 \pm 0.163$ & $0.628 \pm 0.159$
& $35.16 \pm 31.53$ & $21.72 \pm 20.83$ & $34.83 \pm 36.22$ & $30.57 \pm 23.02$ \\

ScribbleVS\cite{VS}
& $0.395 \pm 0.218$ & $0.745 \pm 0.110$ & $0.709 \pm 0.195$ & $0.616 \pm 0.137$
& $85.26 \pm 28.59$ & $31.58 \pm 33.71$ & $52.91 \pm 35.61$ & $56.58 \pm 22.52$ \\

ScribFormer\cite{li2024scribformer}
& $0.388 \pm 0.251$ & $0.598 \pm 0.125$ & $0.701 \pm 0.188$ & $0.562 \pm 0.150$
& $79.38 \pm 34.18$ & $57.54 \pm 33.92$ & $46.84 \pm 38.59$ & $61.25 \pm 25.49$ \\

QMaxViT-Unet+\cite{QMaxvit}
& {\color{blue}$0.705 \pm 0.177$} & {\color{blue}$0.774 \pm 0.071$} & {\color{blue}$0.851 \pm 0.099$} & {\color{blue}$0.776 \pm 0.100$}
& $34.64 \pm 38.08$ & {\color{blue}$10.16 \pm 14.77$} & {\color{blue}$21.16 \pm 28.13$} & {\color{blue}$21.01 \pm 20.85$} \\

Ours
& {\color{red}$0.776 \pm 0.132$} & {\color{red}$0.819 \pm 0.073$} & {\color{red}$0.889 \pm 0.087$} & {\color{red}$0.828 \pm 0.082$}
& {\color{red}$7.78 \pm 15.02$} & {\color{red}$2.09 \pm 1.00$} & {\color{red}$6.85 \pm 17.30$} & {\color{red}$5.57 \pm 7.35$} \\
\bottomrule
\end{tabular}
\end{table*}
\begin{table*}[htp]
\caption{Quantitative comparison of Dice and HD95 with baseline methods on the Prostate test set under the setting of five scribble-annotated training cases. The best and second-best results are highlighted in red and blue, respectively.}
\label{tab:prostate}
\centering
\small
\setlength{\tabcolsep}{6pt}
\begin{tabular}{lcccccc}
\toprule
\multirow{2}{*}{Methods} & \multicolumn{3}{c}{Dice $\uparrow$} & \multicolumn{3}{c}{HD95 (mm) $\downarrow$} \\
\cmidrule(lr){2-4} \cmidrule(lr){5-7}
& PZ & CG & Mean & PZ & CG & Mean \\
\midrule
PCE\cite{PCE}
& $0.165 \pm 0.104$ & $0.356 \pm 0.108$ & $0.260 \pm 0.068$
& $154.53 \pm 14.40$ & $166.68 \pm 22.01$ & $160.61 \pm 15.21$ \\

Cutout\cite{Cutout}
& $0.056 \pm 0.039$ & $0.172 \pm 0.046$ & $0.114 \pm 0.027$
& $152.83 \pm 23.58$ & $151.59 \pm 26.21$ & $152.21 \pm 24.33$ \\

CutMix\cite{Cutmix}
& $0.278 \pm 0.146$ & $0.490 \pm 0.129$ & $0.384 \pm 0.118$
& $153.06 \pm 23.69$ & $150.93 \pm 18.79$ & $151.99 \pm 19.19$ \\

CycleMix\cite{Cyclemix}
& $0.234\pm0.116$ & $0.541\pm0.149$ & $0.388\pm0.111$
& $120.85\pm17.87$ & $115.62\pm43.46$ & $118.23\pm27.24$ \\

CPS\cite{CPS}
& $0.066 \pm 0.048$ & $0.289 \pm 0.124$ & $0.177 \pm 0.061$
& $144.03 \pm 18.10$ & $134.08 \pm 21.14$ & $139.05 \pm 18.11$ \\

MT\cite{MT}
& $0.233 \pm 0.189$ & $0.477 \pm 0.154$ & $0.355 \pm 0.151$
& $80.59 \pm 48.94$ & $39.17 \pm 45.00$ & $59.88 \pm 43.51$ \\

DMPLS\cite{WSL4}
& $0.073 \pm 0.046$ & $0.275 \pm 0.114$ & $0.174 \pm 0.070$
& $153.59 \pm 22.77$ & $163.53 \pm 14.93$ & $158.56 \pm 16.52$ \\

SC-Net\cite{SC-Net}
& $0.149 \pm 0.166$ & $0.305 \pm 0.118$ & $0.227 \pm 0.133$
& $128.76 \pm 35.30$ & $143.10 \pm 31.64$ & $135.93 \pm 27.70$ \\

ScribbleVC\cite{VC}
& {\color{blue}$0.369 \pm 0.158$} & {\color{blue}$0.588 \pm 0.116$} & {\color{blue}$0.478 \pm 0.108$}
& {\color{blue}$33.57 \pm 24.30$} & {\color{blue}$30.80 \pm 40.37$} & {\color{blue}$32.19 \pm 29.43$} \\

ScribbleVS\cite{VS}
& $0.225 \pm 0.156$ & $0.526 \pm 0.138$ & $0.375 \pm 0.115$
& $113.73 \pm 20.77$ & $121.81 \pm 42.14$ & $117.77 \pm 26.77$ \\

ScribFormer\cite{li2024scribformer}
& $0.144 \pm 0.139$ & $0.349 \pm 0.191$ & $0.247 \pm 0.075$
& $143.59 \pm 33.29$ & $161.23 \pm 30.76$ & $152.41 \pm 24.23$ \\

QMaxViT-Unet+\cite{QMaxvit}
& $0.180 \pm 0.073$ & $0.557 \pm 0.116$ & $0.368 \pm 0.065$
& $119.31 \pm 17.34$ & $113.55 \pm 50.99$ & $116.43 \pm 31.66$ \\

Ours
& {\color{red}$0.421 \pm 0.230$} & {\color{red}$0.647 \pm 0.150$} & {\color{red}$0.534 \pm 0.172$}
& {\color{red}$25.59 \pm 24.16$} & {\color{red}$6.59 \pm 3.81$} & {\color{red}$16.09 \pm 12.89$} \\
\bottomrule
\end{tabular}
\end{table*}
\section{Experiments and Results}
\subsection{Datasets}
\textbf{The ACDC dataset} \cite{bernard2018deep} consists of cardiac MRI scans from 100 patients, acquired using scanners with two different magnetic field strengths and spatial resolutions. The segmentation targets include the left ventricle (LV), myocardium (MYO), and right ventricle (RV). Following~\cite{MAAG,VC}, we adopt the provided scribble annotations and partition the dataset into training, validation, and test sets according to the standard MAAGfold protocol, with a ratio of 70/15/15. To evaluate the effectiveness of the proposed method under a few-shot scribble-supervised setting, only the first five cases in the training set are used as scribble-labeled samples, while the remaining 65 training cases are treated as unlabeled training data.

\textbf{The Prostate dataset}~\cite{clark2013cancer} is derived from the T2-weighted MRI scans released in the ISBI 2013 Prostate MRI Challenge, containing 80 prostate volumes in total. The segmentation targets are the central gland (CG) and peripheral zone (PZ). We use the scribble annotations provided by \cite{WSL4,luo2022word}, and split the dataset in ascending order of sample IDs into 55 training cases, 10 validation cases, and 15 test cases. To construct the few-shot scribble-supervised setting, we retain only the scribble annotations of the first five training cases as labeled data, while treating all remaining training cases as unlabeled samples during training.
\subsection{Implementation Details}
Our method is implemented in PyTorch and trained on a single NVIDIA RTX A6000 GPU. For both the ACDC and Prostate datasets, each slice is first intensity-normalized to the range of $[0,1]$. To improve data diversity, random rotation and random flipping are employed as data augmentation during training. All augmented images are resized to $256 \times 256$ before being fed into the network.

We use U-Net~\cite{ronneberger2015u} as the segmentation backbone and adopt the shallow CNN architecture of SSN~\cite{jampani2018superpixel} for the superpixel model, with a feature dimensionality of 20 and 100 superpixels. The framework is optimized with AdamW~\cite{loshchilov2018decoupled} using an initial learning rate of $1\times10^{-3}$ and a cosine annealing schedule. Before bi-level collaborative training, the segmentation and superpixel models are separately pretrained for 5{,}000 iterations to stabilize optimization. Specifically, the segmentation model is initialized with Mean Teacher, using a pseudo-label confidence threshold of $\tau=0.9$ and a moving average coefficient of $\alpha=0.99$. The superpixel model is pretrained without the semantic reconstruction loss from segmentation feedback, with $\gamma=1.8$ and $\beta=2.0$. During bi-level optimization, the lower- and upper-level update steps are set to $N_l=50$ and $N_u=5$, respectively. The maximum number of training iterations is $T=20{,}000$, and each mini-batch contains 12 slices, including 6 scribble-labeled and 6 unlabeled slices. In the upper-level objective, the loss weights are set to $\lambda_{\mathrm{low}}=0.5$, $\lambda_{\mathrm{sem}}=1$, and $\lambda_c=10^{-4}$. For the spatial-prior-guided filtering strategy, we set $\gamma_{\mathrm{center}}=6$ and $\sigma=0.3$. The same hyperparameter settings are used across all datasets.
\subsection{Comparison and Evaluation}
To comprehensively evaluate the proposed method, we compare it with a diverse set of representative baselines, including a basic method based on partial cross-entropy loss (PCE)~\cite{PCE}, data augmentation strategies, semi-supervised methods, and scribble-supervised methods. Specifically, the augmentation-based baselines include the classical Cutout \cite{Cutout} and CutMix \cite{Cutmix}, as well as CycleMix \cite{Cyclemix}, which is specifically designed for scribble supervision. The semi-supervised baselines include two canonical frameworks, CPS \cite{CPS} and MT \cite{MT}. The scribble-supervised baselines include DMPLS \cite{WSL4}, SC-Net \cite{SC-Net}, ScribbleVC \cite{VC}, ScribbleVS \cite{VS}, ScribFormer \cite{li2024scribformer} and QMaxViT-Unet+ \cite{QMaxvit}. For fair comparison, we evaluate both the scribble-only and semi-supervised training settings for baselines involving pseudo-label learning, and report the better result.

Table~\ref{tab:acdc} reports the results on ACDC under the five-case scribble-supervised setting. Our method achieves the best overall performance, outperforming the second-best baseline, QMaxViT-Unet+, by 5.2 percentage points in Mean Dice and 15.44 mm in Mean HD95. The advantage is further validated on the more challenging Prostate dataset in Table~\ref{tab:prostate}, where our method improves Mean Dice by 5.6 percentage points over ScribbleVC and reduces Mean HD95 by 16.10 mm. These results demonstrate the effectiveness of the proposed BiSCL in improving both region recovery and boundary delineation under sparse supervision and limited annotated samples.

\begin{figure}[h]
    \centering
    \includegraphics[width=1.0\linewidth,height=3cm]{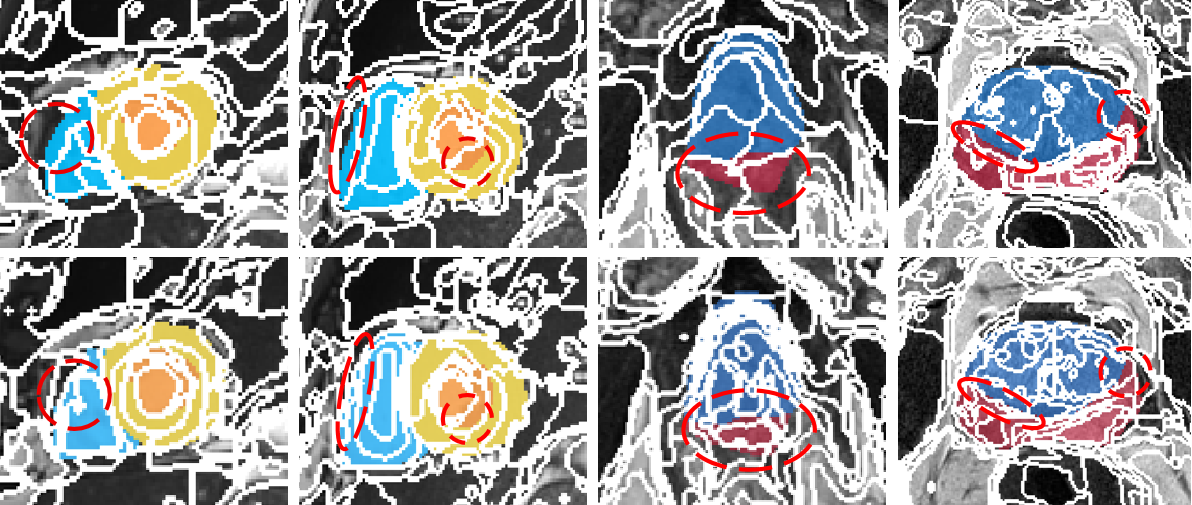}
    \caption{Visual comparison of superpixel partitioning before (top row) and after (bottom row) bi-level collaborative learning, where the colored masks denote the segmentation ground truth.}
    \label{fig:ssn}
\end{figure}

The qualitative comparisons further support the above observations. As shown in Fig.~\ref{fig:acdc}, methods such as DMPLS and SC-Net are prone to evident foreground-background confusion, while MT, CycleMix, and several recent scribble-based methods still suffer from incomplete region recovery, boundary deviations, or unstable predictions in challenging regions. In contrast, our method yields predictions that are visually closer to the ground truth, with more compact target regions, cleaner inter-class boundaries, and fewer scattered false positives. This advantage stems from the bidirectional interaction between the upper and lower levels, which enables the superpixel model to dynamically provide structural priors better aligned with the lower-level segmentation task, thereby generating reliable dense pseudo-labels to alleviate the dual challenges of sparse supervision and limited annotated samples. To further substantiate this point, Fig.~\ref{fig:ssn} presents a comparison of the superpixel maps before and after bi-level collaborative training. Before collaboration, the superpixel model mainly partitions the image according to low-level visual cues, and thus tends to produce regions that cross segmentation semantics. After receiving semantic feedback from the lower-level segmentation model during collaborative learning, it generates superpixels that are better aligned with segmentation semantics.
\subsection{Ablation Study}

\begin{figure}[h]
    \centering
    \includegraphics[width=1.0\linewidth,height=3cm]{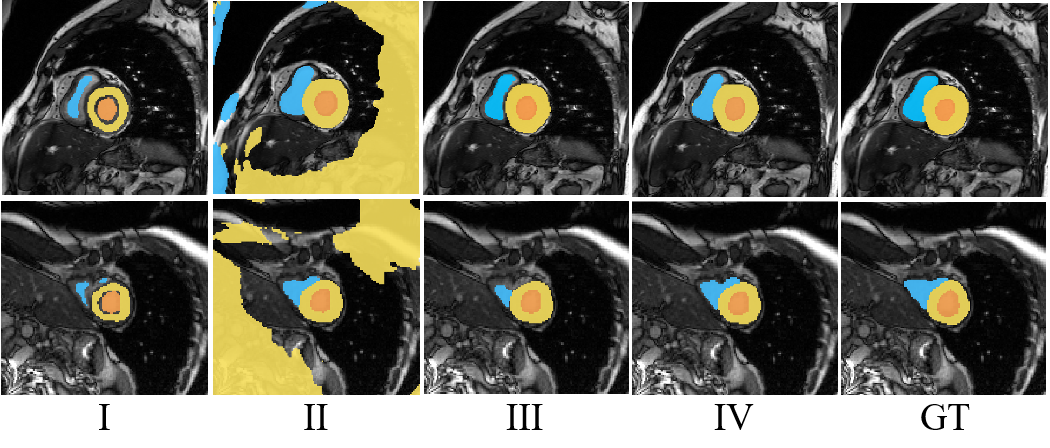}
    \caption{Qualitative comparison of different ablation variants.}
    \label{fig:Ablation}
\end{figure}

Table~\ref{tab:ablation} reports the ablation results of the key components of the proposed BiSCL method on the ACDC dataset. Starting from the MT baseline, directly introducing Superpixel Propagation causes a clear performance drop, with the Mean Dice decreasing from 0.599 to 0.311 and the Mean HD95 increasing from 53.78 to 93.20. This indicates that under the few-shot scribble-supervised setting, sparse supervision and limited annotated samples make the model highly vulnerable to pseudo-label noise, such that unconstrained superpixel diffusion tends to amplify erroneous predictions rather than improve region completion, as illustrated in Fig.~\ref{fig:Ablation}.

After further incorporating the proposed spatial-prior-guided filtering strategy, the performance improves substantially, boosting the Mean Dice to 0.774 while reducing the Mean HD95 to 22.87. This confirms that the proposed filtering mechanism can effectively suppress unreliable diffusion and make the superpixel prior practically beneficial for both region recovery and boundary delineation. Finally, further introducing the bi-level collaborative learning strategy yields the best performance, achieving a Mean Dice of 0.828 and a Mean HD95 of 5.57. This demonstrates that, instead of relying on static superpixel priors, the proposed collaborative optimization between the superpixel and segmentation models enables the learned superpixels to better align with the segmentation objective, thereby producing more reliable pseudo-labels and further improving segmentation quality.
\begin{table}[h]
\centering
\caption{Ablation study of different components on the ACDC dataset.}
\label{tab:ablation}
\setlength{\tabcolsep}{1.8pt}
\small
\begin{tabular}{c|cccc|cc}
\toprule
ID & MT & \makecell{Superpixel\\Propagation} & \makecell{Spatial-Prior\\Filtering} & \makecell{Bilevel\\Training} & Mean Dice & Mean HD95 \\
\midrule
I & \checkmark &  &  &  & $0.599 \pm 0.106$ & $53.78 \pm 25.43$ \\
II &\checkmark & \checkmark &  &  & $0.311 \pm 0.146$ & $93.20 \pm 14.12$ \\
III &\checkmark & \checkmark & \checkmark &  & $0.774 \pm 0.115$ & $22.87 \pm 23.07$ \\
IV &\checkmark & \checkmark & \checkmark & \checkmark & $\mathbf{0.828 \pm 0.082}$ & $\mathbf{5.57 \pm 7.35}$ \\
\bottomrule
\end{tabular}
\end{table}
\subsection{Hyperparameter Sensitivity Analysis}
We further analyze the sensitivity of the parameters in the proposed spatial-prior-guided filtering strategy on ACDC under the few-shot scribble-supervised setting. As shown in Table~\ref{tab:param_sensitivity}, the combination of $\sigma=0.3$ and $\gamma_{\text{center}}=6$ achieves the best performance. When either parameter is moderately perturbed around this default setting, the overall performance remains relatively stable and competitive across different settings, indicating that the proposed strategy is reasonably robust to parameter variations. Overall, the results suggest that a moderate parameter setting strikes a better balance between preserving valid region expansion and suppressing noisy propagation.
\begin{table}[H]
\centering
\caption{Sensitivity analysis of the hyperparameters in the spatial-prior-guided filtering strategy on the ACDC dataset.}
\label{tab:param_sensitivity}
\setlength{\tabcolsep}{2.8pt}
\small
\begin{tabular}{ccc|ccc}
\toprule
\multicolumn{3}{c|}{Varying $\sigma$ ($\gamma_{\text{center}}=6$)} 
& \multicolumn{3}{c}{Varying $\gamma_{\text{center}}$ ($\sigma=0.3$)} \\
\cmidrule(r){1-3} \cmidrule(l){4-6}
$\sigma$ & Mean Dice & Mean HD95 & $\gamma_{\text{center}}$ & Mean Dice & Mean HD95 \\
\midrule
0.2 & $0.809 \pm 0.094$ & $10.19 \pm 19.10$ & 5 & $0.810 \pm 0.098$ & $8.96 \pm 10.39$ \\
0.3 & $\mathbf{0.828 \pm 0.082}$ & $\mathbf{5.57 \pm 7.35}$ & 6 & $\mathbf{0.828 \pm 0.082}$ & $\mathbf{5.57 \pm 7.35}$ \\
0.4 & $0.823 \pm 0.088$ & $10.60 \pm 17.39$ & 7 & $0.826 \pm 0.088$ & $9.54 \pm 17.87$ \\
\bottomrule
\end{tabular}
\end{table}

\section{Conclusion}
This paper addresses the practical medical imaging scenario where sparse supervision and limited annotated samples coexist by proposing a bi-level collaborative learning framework for few-shot scribble-supervised medical image segmentation. Specifically, an upper-level superpixel network is introduced to provide region-structural priors for lower-level segmentation, supporting pseudo-label diffusion, while a spatial-prior-guided filtering mechanism is designed to generate more reliable dense supervisory signals. Meanwhile, the segmentation semantics learned at the lower level are fed back to the upper level to encourage the superpixel network to learn structural representations better aligned with the segmentation task. Through such bidirectional interaction and collaborative optimization, the proposed framework effectively exploits the structural priors encoded in superpixels to facilitate segmentation learning. Experiments on the ACDC and Prostate datasets show that, with only five scribble-annotated cases, the proposed method consistently outperforms existing methods.

\bibliographystyle{ACM-Reference-Format}
\bibliography{samples/bilevel}










\end{document}